%% file: main.tex
\documentclass[10pt, conference, compsocconf]{IEEEtran}
\usepackage{cite}

\usepackage[pdftex]{graphicx}

\usepackage[cmex10]{amsmath}
\usepackage{algorithmic}

\usepackage{array}

\usepackage{mdwmath}
\usepackage{mdwtab}

\usepackage[tight,footnotesize]{subfigure}

\usepackage{graphicx}
\usepackage{caption}
\usepackage[font=footnotesize]{subfig}

\usepackage{url}
\usepackage{tabularray}
\usepackage{bm}

\newcommand{\set}{\mathcal}

\usepackage{acronym}
\acrodef{CSS}{Continual Semantic Segmentation}
\acrodef{CNN}{Convolutional Neural Network}
\acrodef{BACS}{Background Aware Continual Semantic Segmentation}
\acrodef{KD}{Knowledge Distillation}
\acrodef{DER}{Dark Experience Replay}
\acrodef{Dec}{Transformer Decoder}
\acrodef{GEM}{Gradient Episodic Memory}
\acrodef{LwF}{Learning Without Forgetting}
\acrodef{SSUL}{Semantic Segmentation with Unknown Label}
\acrodef{MKD}{Masked Knowledge Distillation}

\usepackage{float}
\usepackage{amssymb}
\usepackage{pifont}
\usepackage{booktabs} 
\newcommand{\xmark}{\ding{55}}
\usepackage{wrapfig}
\usepackage{tabularx} 
\usepackage{cleveref}

\begin{document}
%
\title{BACS: Background Aware Continual Semantic Segmentation}


\author{
\IEEEauthorblockN{Mostafa ElAraby}
\IEEEauthorblockA{DIRO, Mila - Quebec AI Institute\\
Universit\'e de Montr\'eal\\
Montreal, Canada \\
\texttt{\{elarabim\}@mila.quebec}} 
\and 

\IEEEauthorblockN{Ali Harakeh}
\IEEEauthorblockA{DIRO, Mila - Quebec AI Institute\\
Universit\'e de Montr\'eal\\
Montreal, Canada \\ }

\and
\IEEEauthorblockN{Liam Paull }
\IEEEauthorblockA{CIFAR AI Chair \\ 
DIRO, Mila - Quebec AI Institute\\
Universit\'e de Montr\'eal\\
Montreal, Canada \\ }
}

\maketitle

\begin{abstract}
\input{sections/00-abstract}
\end{abstract}


\section{Introduction} \label{sec:intro}
\input{sections/01-introduction}

\section{Related Work} \label{sec:related}
\input{sections/02-related.tex}

\section{Proposed Method}\label{sec:method}
\input{sections/03-method}

\section{Experiments and Results} \label{sec:experiments}
\input{sections/04-experiments}

\section{Conclusion} \label{sec:conclusion}
\input{sections/05-conclusion}


\bibliography{main}   
\bibliographystyle{IEEEtran}
\end{document}

%% file: sections/00-abstract.tex
Semantic segmentation plays a crucial role in enabling comprehensive scene understanding for robotic systems. However, generating annotations is challenging, requiring labels for every pixel in an image. In scenarios like autonomous driving, there's a need to progressively incorporate new classes as the operating environment of the deployed agent becomes more complex. For enhanced annotation efficiency, ideally, only pixels belonging to new classes would be annotated. This approach is known as \ac{CSS}. Besides the common problem of classical catastrophic forgetting in the continual learning setting, \ac{CSS} suffers from the inherent ambiguity of the background, a phenomenon we refer to as the ``background shift'', since pixels labeled as background could correspond to future classes (forward background shift) or previous classes (backward background shift). As a result, continual learning approaches tend to fail. This paper proposes a Backward Background Shift Detector (BACS) to detect previously observed classes based on their distance in the latent space from the foreground centroids of previous steps.
Moreover, we propose a modified version of the cross-entropy loss function, incorporating the BACS detector to down-weight background pixels associated with formerly observed classes. To combat catastrophic forgetting, we employ masked feature distillation alongside dark experience replay. Additionally, our approach includes a transformer decoder capable of adjusting to new classes without necessitating an additional classification head. We validate BACS's superior performance over existing state-of-the-art methods on standard \ac{CSS} benchmarks.

%% file: sections/01-introduction.tex
A typical assumption in training deep neural network models is the availability of the entire dataset at training time. However, in many applications, incrementally learning a new stream of classes without forgetting previously-learned knowledge is required for achieving human-level intelligence. When we fine-tune a model on a new set of classes, it may forget previously acquired knowledge (catastrophic forgetting)~\cite{robins1995catastrophic, french1999catastrophic, thrun1998lifelong}. In recent years, \acs{CNN} have shown tremendous progress on many computer vision tasks, including semantic segmentation, with all classes learned jointly in a single shot~\cite{chen2018encoder,zhang2020resnest, yurtkulu2019semantic}. However, during deployment on a robot, for example, novel classes may emerge which necessitate updating the model. For that purpose, \acf{CSS}~\cite{ozdemir2018learn, ozdemir2019extending,michieli2019incremental,cermelli2020modeling} learn new classes incrementally without a need for retraining on data from previous classes.

\begin{figure}[ht]
    \centering
    \includegraphics[width=\columnwidth]{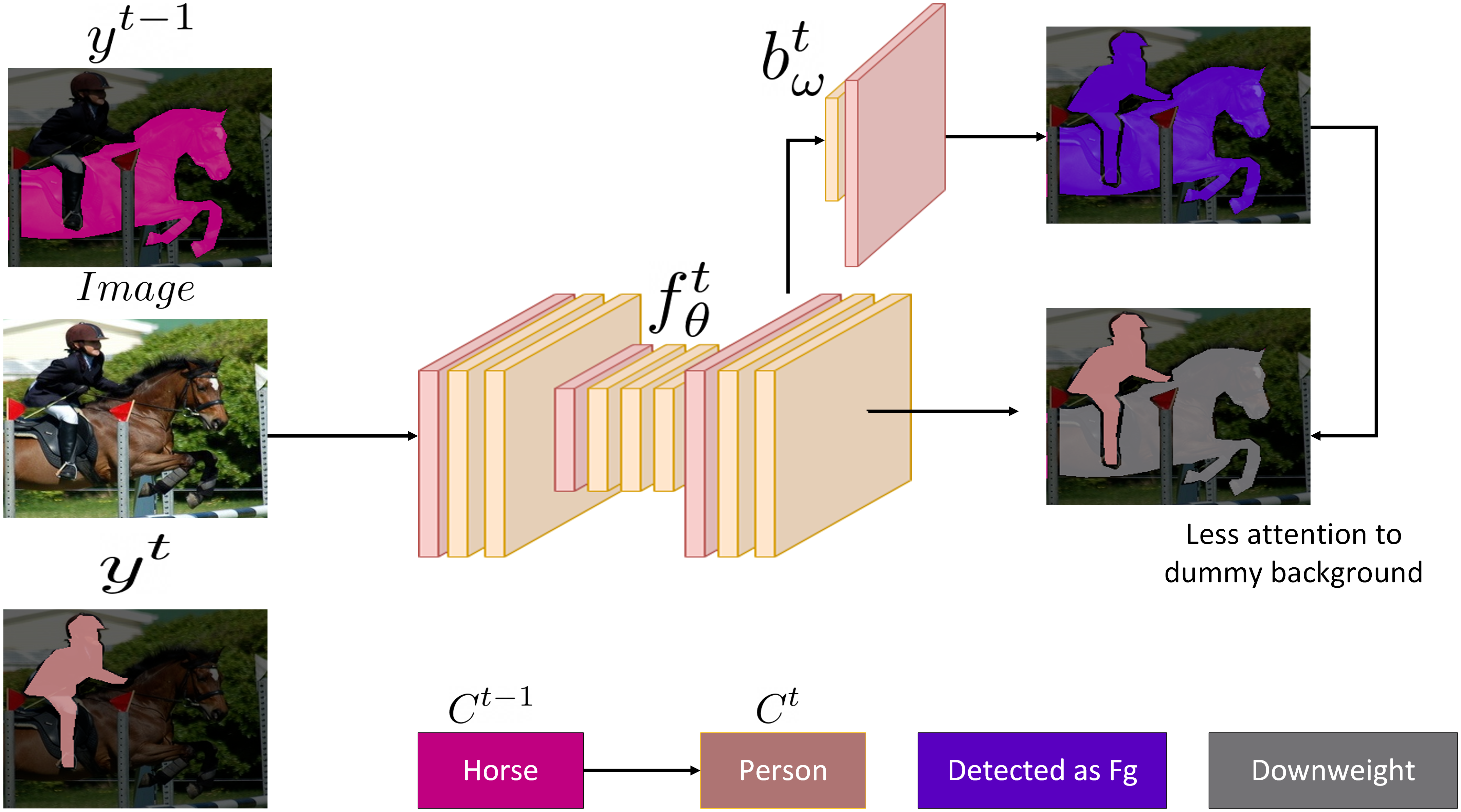}
    \caption{\ac{BACS} framework overview. A backward background detector \(b_{\omega}^t\) down weights background pixels detected that have appeared as classes in previous steps to avoid the collapse of old classes into the background. }
    \label{fig:csl}
\end{figure}

One of the main challenges of class-incremental learning is the catastrophic forgetting of previously observed data points distributions while learning distributions with data from new classes.
Existing class incremental learning literature creates a trade-off between learning new knowledge (plasticity) and keeping previously acquired knowledge (stability) \cite{zilly_NeurIPS_2021}.
\Ac{CSS} poses more challenges exacerbating the catastrophic forgetting phenomenon.
In \ac{CSS}, annotations are present only for new classes.
By default, everything else is labeled as background to make it easier and more realistic during the data collection and annotation.
The collapse of old and future classes into the background label results in a shift in the background distribution during training which exacerbate further the forgetting of old classes.
We distinguish between two types of background shifts, forward and backward.
In the forward background shift, the distribution of the background class is shifting towards the current step's new classes.
On the other hand, in the backward background shift, the distribution of old classes is moving to the current step's background class. The collapse of old and future classes into the background class causes a misalignment of features collected from previous steps and exacerbates catastrophic forgetting.

Recent works attempt to address both background shift and catastrophic forgetting using a custom \ac{KD}~\cite{hinton2015distilling} technique used extensively in \ac{CSS}.
\ac{CSS} literature~\cite{douillard2021tackling,douillard2021plop, michieli2019incremental, cermelli2020modeling} focuses on teacher-student \ac{KD} in all classes while taking into account the background shift.
However, the more we constrain our student network, the better the stability at the expense of the network's plasticity.
Moreover, the network's performance on new classes will rely heavily on the order and number of classes used in the initial step, making it harder to start the learning process from a few classes.
For the background shift problem, current literature relies on the pseudo-labeling of existing data to detect previous classes, causing potential overfitting of errors coming from old classes known as the problem of confirmation bias in pseudo-labeling~\cite{arazo2020pseudo}.

Furthermore, adding new classes in each step, we need to update our architecture to accommodate new classes.
\ac{CSS} baselines use a separate classification head where we initialize a new head per new upcoming set of classes without knowing the step number during inference (class-incremental setup).
Random initialization of new heads could provoke a misalignment among background class features learned by the previous model~\cite{michieli2019incremental}, affecting the stability of the network if not initialized using the background class weights.

Our paper first identifies the drawback of using pseudo-labeling and output \ac{KD} during the training and its effect on the plasticity of the network. First, we introduce a backward background detector, BACS, network connected to our latent space representation to detect if the pixel is a ``true'' background or corresponds to an old class from any previously observed step, as shown in \cref{fig:csl}. Next, we incorporate the output of BACS into our loss function to mitigate the effect of the background shift.
Moreover, we propose a \acf{MKD} on the penultimate layer's features that only focus on background pixels detected as foreground.
Lastly, we use a transformer decoder as the output classification layer to avoid the initialization trick of new heads and to make the model use fewer parameters than the multi-classifier setup.

In summary, our contributions are the following:
\begin{itemize}
    \item We propose a  backward background shift detector trained to detect the background and foreground of each step based on the distance from saved foreground prototypes.
    \item We introduce a variation of the cross-entropy loss function that uses the output of \ac{BACS} during training to mitigate catastrophic forgetting.
    \item We propose a feature-based \ac{MKD} that uses the information from the down-weighted background pixels.
    \item Finally, we suggest using a transformer-like decoder to avoid both initializing new heads and copying the weights of the background class~\cite{michieli2019incremental}.
\end{itemize}

We demonstrate the effectiveness of \ac{BACS} in addressing the background shift problem and handling long steps while improving by a large margin both the plasticity and stability of the network compared to existing baselines~\cite{cermelli2020modeling,douillard2021plop,maracani2021recall,michieli2021continual}.
Moreover, our method adds new classes by simply appending a single parameter without perturbing the background class's feature space, resulting in better stability. Finally, we perform a set of ablation studies to show the effect of each proposed component on our results.

%% file: sections/02-related.tex
This section will summarize the most important related work in semantic segmentation, continual learning, and continual semantic segmentation \ac{CSS}.

\subsection{Semantic Segmentation}

Early methods~\cite{krahenbuhl2011efficient, zheng2015conditional, arnab2016higher} for semantic segmentation relied on classifying patches of input images, then refining the predictions based on the context.
Later, fully convolutional networks (FCN)~\cite{long2015fully,goyal2017multi} enable the substitution of fully connected layers with convolutional ones to produce spatial maps. Compared with traditional methods, the FCN model improved various semantic segmentation tasks achieving end-to-end semantic segmentation. The U-Net~\cite{ronneberger2015u} improved over the FCN by learning context and spatial information using convolutional upsampling.
Current methods~\cite{badrinarayanan2017segnet, hariharan2015hypercolumns,lin2018multi,lin2017refinenet,long2015fully,noh2015learning, peng2017large,tian2019decoders} focus on learning multi-scale feature aggregation from existing pre-trained convolutional networks.
Recent methods~\cite{chen2016attention,ding2018context, fu2019dual, li2019expectation} relied on the attention score to learn the connections between image contexts.
Another set of methods attempted to fuse various receptive fields of view using atrous convolutions~\cite{chen2018encoder,niu2018deeplab} and spatial pyramids with dilated convolutions~\cite{mehta2018espnet, yang2018denseaspp} in an encoder-decoder setup.
Following the success of transformers in natural language processing~\cite{vaswani2017attention}, an adapted transformer architecture has become prevalent for computer vision tasks~\cite{dosovitskiy2020image,wang2021end,zeng2020learning,zheng2021rethinking}, showing an improvement over existing architectures.
Existing transformer architectures in semantic segmentation either focus on multi-scale feature fusion~\cite{chen2021crossvit, wang2021pyramid, xie2021segformer,zhang2020feature} or contextual feature aggregation~\cite{liu2021swin, strudel2021segmenter}.
In our work, we extend DeepLabV3~\cite{chen2018encoder} to have a dynamically extendable decoder similar to the decoder of Segformer~\cite{xie2021segformer}. Our proposed decoder enables us to extend our model with novel classes with low memory and compute requirements.

\subsection{Continual Learning}

Continual learning focuses on learning from a sequential data stream to gradually extend acquired knowledge without forgetting old knowledge. Data can stem from different domains (covariate shift) or different tasks.
In some literature, continual learning is also called lifelong learning~\cite{chen2018lifelong,aljundi2017expert,chaudhry2018efficient,parisi2019continual}, sequential learning~\cite{aljundi2018selfless,mccloskey1989catastrophic,shin2017continual} or incremental learning~\cite{aljundi2018memory,chaudhry2018riemannian, gepperth2016bio,rebuffi2017icarl}.
The major challenge in continual learning is learning new tasks without suffering from catastrophic forgetting.
This problem originates from the plasticity-stability dilemma~\cite{dobson1984reviews} in deep learning systems. Plasticity refers to the ability to integrate new knowledge and stability in retaining previous knowledge while learning new ones.
In \ac{CSS}, we focus on the task incremental learning setting since incremental tasks share the same background class. In incremental learning, we can distinguish three methods: replay-based, regularization-based, and Parameter isolation methods.

Replay methods rely on storing samples from old tasks or generating pseudo-samples with a generative model, then replaying them while learning new tasks to avoid forgetting.
The end-to-end incremental learning~\cite{castro2018end} relies on keeping a balanced memory of previously seen classes and merges them with the new task data in an end-to-end learning fashion. \acf{DER}~\cite{buzzega2020dark} relies on replaying previous tasks' data saved along with their logits from earlier tasks, thus matching the network's output with its past through the optimization trajectory.
\acf{GEM}~\cite{lopez2017gradient} uses exemplars to solve a constrained optimization problem that chooses gradient updates in the direction of learning new tasks while retaining knowledge of previously seen classes. We can consider replay data as some low-resource training with few data points.
In our proposed framework, we mitigate the catastrophic forgetting using \acf{DER}~\cite{buzzega2020dark} as it provides a solid baseline empirically shown to converge to flatter minima compared to vanilla experience replay.

Regularization-based approaches for continual learning avoid saving samples from previous tasks, prioritizing privacy, and reducing the memory used to store previous samples.
Instead, these approaches add a regularization term to consolidate previous knowledge while learning new classes. The initial idea was to save the output of an earlier task model given a new input image to alleviate catastrophic forgetting~\cite{silver2002task}.
That same approach has been re-introduced by \acf{LwF}~\cite{li2017learning} using a \acf{KD} loss while training on new tasks to preserve the decision boundary of the neural network.
However, these methods are vulnerable to domain shift between tasks~\cite{aljundi2017expert}, as they tend to keep the feature space near its counterpart trained on the previous task.
For that reason, shallow autoencoders are used to constrain task features in their corresponding learned low dimensional space~\cite{rannen2017encoder}, thus reducing the negative effect of domain shift between tasks.

In our proposed framework, we use a soft teacher-student knowledge distillation on the penultimate layer that only focuses on features belonging to old classes without affecting the plasticity of the network.

\subsection{Continual Semantic Segmentation}

In continual semantic segmentation, the background class might include pixels associated with previously observed classes from earlier steps, as in \cref{fig:csl}, exacerbating catastrophic forgetting. Most of the existing work~\cite{douillard2021plop,michieli2019incremental,cermelli2020modeling,michieli2021continual,baek2022decomposed,phan2022class,yang2022uncertainty} keeps the old network from the previous step and uses \ac{KD} to keep old step information. Using pseudo-labeling and a teacher-student approach makes starting with a few classes harder, as the optimization procedure tends to keep the network close to its weights in the initial step.
Pod distillation loss~\cite{douillard2021plop} regularizes the current step's network output with the previous one at each output layer and discards uncertain pseudo-labels. Later Pod distillation was extended using an object replay buffer to support long sequences~\cite{douillard2021tackling}.
Matching class representation prototypes and repulsing different classes' representation~\cite{michieli2021continual} improved the representation space and reduced the effect of catastrophic forgetting compared with the classic knowledge distillation loss. \cite{zhang2022representation} proposed a custom convolutional layer that fuses weights from the previous step layer with the current one to reduce the effect of catastrophic forgetting.
Another set of methods~\cite{maracani2021recall,huang2021half,yan2021framework} has developed several replay-based methods for \ac{CSS}.
Finally, a separate feature extractor is created per step and frozen at the end of the step, thus reducing the catastrophic forgetting to its minimum while using binary cross-entropy loss to avoid the background shift~\cite{cha2021ssul}.

Our proposed approach keeps a small buffer of examples and their corresponding logits from previous steps. We use BACS to differentiate between the background and old classes instead of pseudo-labeling that would accumulate errors. In the subsequent section, we explain how we detect and disentangle the background class from previously observed foreground classes in continual semantic segmentation.

%% file: sections/03-method.tex
\subsection{Notation and Problem Setting}

In \acf{CSS}, we observe a set of incremental classes \( t= 1 \ldots T\), each having a dataset \(  D^{t} \). In each incremental step \(t\), the model has to learn a set of classes \(C^t\) using dataset \(D^t\) consisting of a set of 2D images \(x^t\) of size \(N=H \times W\) and 2D ground truth per pixel labels \(y^t_i \in C^t \; \forall i=1\ldots N\). 
The label space of each step \(t\) consists of new classes \(C^t\) and a dummy background class \(c_{bg}\), which is associated with every pixel that was not labeled. 
We assume there is no intersection between label spaces \(C\) of various steps, resulting in a dummy background class \(c_{bg}\) possibly containing future and old classes. 
We denote the collapse of old classes into the background class by \textit{backward background shift} and the collapse of new classes by \textit{forward background shift}. 
We follow the same set of scenarios considered in~\cite{michieli2019incremental, douillard2021plop} grouped into three modes sequential, disjoint, and overlap. 
The sequential mode keeps the ground truth of all labels $C^{1:t}\triangleq \bigcup_{t^\prime = 1}^t C^{t^\prime}$ observed so far. On the other hand, the disjoint mode keeps only the ground truth of the current step \(C^t\) while having the backward background shift. The third and final mode is overlap that uses all images having at least a class from \(C^t\) with everything else as \(c_{bg}\) and thus suffers from both backward and forward background shift. In our work, we consider the overlap mode, which is the most realistic and challenging.

\begin{figure*}[ht]
    \centering
    \includegraphics[width=\linewidth,trim={6 6 6 6}, clip]{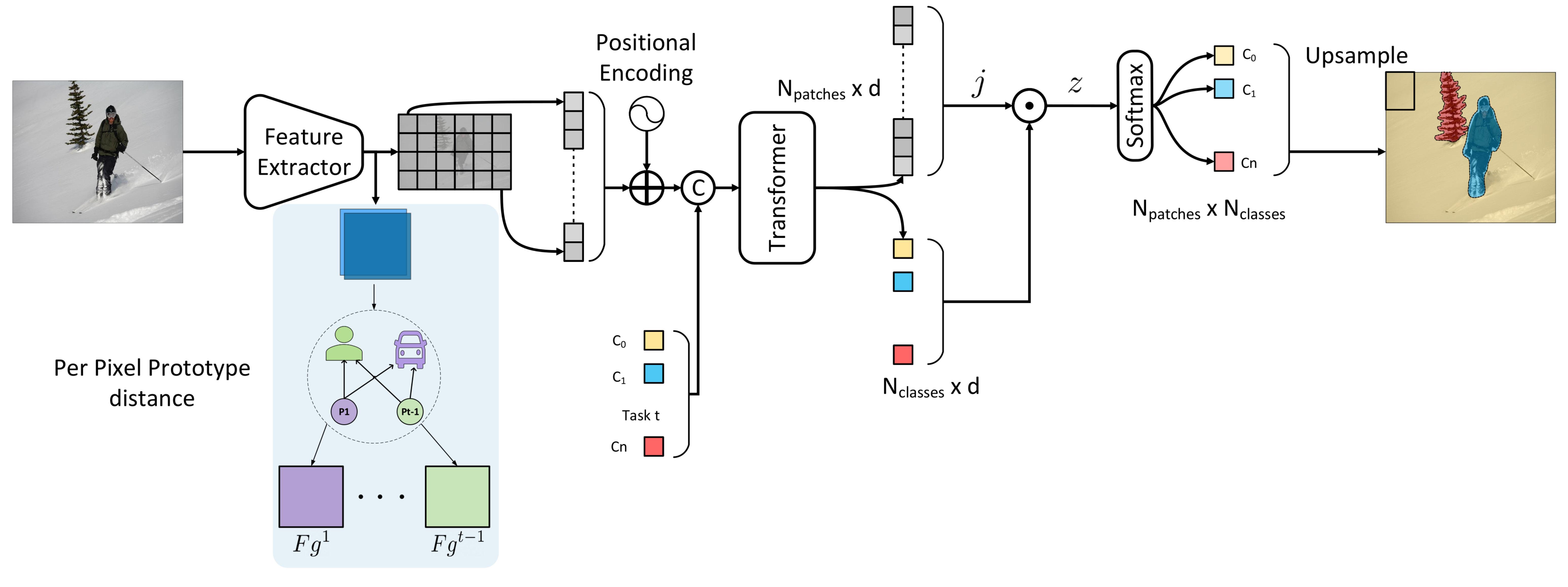}
    \caption{Our continual learning framework BACS consists of the backward background detector, shown in blue, and a transformer decoder. The backward background detector compares the latent space of each pixel with a per-step centroid to detect the foreground. The maximum output probability of all heads, \( \max{Fg^{1:t-1}}\), is used to reduce the emphasis on pixels that belong to old classes collapsing to the background class in step \(t\) ground truth. Next, the transformer decoder allows the addition of new classes by initializing new class tokens.}
    \label{fig:framework}
\end{figure*}

We define our semantic segmentation model \(f_{\theta}^t\) to be a mapping from input image \(x^t\) to per-pixel predictions, as shown in \cref{fig:framework}. 
We denote the free logits before the softmax layer for input $x^t$ by \( z^t \in \mathbb{R}^{N \times \lvert C^{1:t}  \rvert}\) ($z^t[i,c]$ denotes the logit value for pixel $i$ that corresponds to class $c$), and the hidden penultimate layer per pixel features for input $x^t$ by \( j^t  \in \mathbb{R}^{N \times \lvert d \rvert} \) where \(d\) is the size of the decoder's hidden representation. 
Our model \(f_{\theta}^t\) consists of a convolutional feature extractor \(h^t\) and a decoder \( g^t\)\footnote{Convention \(f_{\theta}^t\) means model trained at the end of step \(t\). Still, we use a single network trained at the end of all steps during inference. On the other hand, output heads are separate per step in the conventional multi-convolutional setup~\cite{cermelli2020modeling}}. 
Existing baselines' architecture uses a decoder \( g^t\) consisting of another feature extractor and a separate classifier per step \(t\); thus, we do not need to know the step id during inference. 
However, due to the forward background shift, random initialization of new classifiers would perturb the feature space from the background to the new classes. 
For that reason, recent methods~\cite{michieli2019incremental, michieli2021continual,douillard2021plop, cha2021ssul, douillard2021tackling, maracani2021recall} used \textit{unbiased initialization trick } that uses the background class weights to initialize new heads in a way that incentivize the model to predict new classes as the background class. Furthermore, we denote \( \mathcal{M} \) as the exemplar balanced memory consisting of a small set of samples from previous steps \(D^{1:t-1}\) along with their corresponding logits at its step \(t\),  and use it as a replay for the current step.

CSS faces the same inherent challenges of continual learning; the model forgets old classes while learning new ones. Without a forward or backward background shift, we can adapt continual learning methods to achieve acceptable results in the sequential mode. However, CSS comes with two additional challenges. The first one is the background shift which we split into forward and backward for simplicity. The second is the initialization of new heads in a forward background shift where the network might observe future classes as a background class. Existing baselines~\cite{douillard2021plop, cermelli2020modeling, michieli2019incremental, michieli2021continual} solves the first challenge by either pseudo-labeling using the previous step network \(f_{\theta}^{t-1}\) to detect backward background shift or to train separate feature extractors and classifiers using binary cross-entropy while freezing old steps' heads~\cite{cha2021ssul}. 
For the forward background shift, the parameters of new classifiers are initialized precisely as the background class to reduce the feature space perturbation, thus decreasing its counter-effect on the background class.

\subsection{Backward Background Shift Detector}

This section proposes an effective method to tackle background shift in \ac{CSS}. 
First, we introduce our backward background shift detector \( b_{\omega} \)  that follows a Siamese network architecture~\cite{koch2015siamese} comparing input pixel representation with each foreground representation saved as a running mean per step \(t\) (prototype) \(P^t\).
 The backward background shift detector maps the encoder's feature space \( h^t(x^t)\) to multi-label classification detecting whether a pixel belongs to the foreground of any of the classes previously observed using a binary probability \(Fg^{t} \in \mathbb{R}^{N \times \lvert t \rvert }\). 
 The higher the probability \(Fg^t\) at step \(t\), the higher the probability of being a foreground at step \(t\).
The proposed backward background detector \( b_{\omega} \)  consists of a convolutional feature projector \( K \), trained only in the initial step, followed by a set of \( 1 \times 1  \) convolutional filters  \( \phi_{\Phi}^{t}  \)~\footnote{here \(\phi_{\Phi}^{t}\) denotes a separate convolutional filter per each step}. 
We maintain prototype \(P^t\) as a representative of the foreground classes in step \(t\) by computing a running mean of the projection's network output \( K( h^t(x^t))  \) during the training phase.

Following the analysis from~\cite{ramasesh2020anatomy}, deeper layers contribute the most to forgetting, and thus we need to only train the projection network \(K\) on the first step. 
For subsequent steps, we only train step \(t\)'s output convolutional filter \( \phi_{\Phi}^{t} \) to disentangle \(Fg^t\) from anything else since we only have the ground truth \(C^{t}\), and everything else is collapsed to \(c_{bg}\).  
For previous step heads, we freeze their corresponding output heads \( \phi_{\Phi}^{1:t-1} \) and rely on the prototypes \(P^t\) to retain previous steps' knowledge in detecting their corresponding foreground. 
We use the binary focal loss~\cite{lin2017focal} \(\ell^{\omega^t}_{bacs}(x^t,y^t)\) to train each detection head with the projection network \(K\) only trained on the first step.
Hence, our detector can identify backward background shift based on the maximum logit value from each trained head. Next, we introduce our loss function that uses the backward background shift detector.

\subsection{Backward Background Shift Aware Loss}
With the background shift, a conventional cross-entropy loss would exacerbate catastrophic forgetting as the model learns to associate old classes with the dummy background. 
Hence, we introduce a backward background shift aware loss function that uses the output of our background detector heads to guide the training. 
Our new loss, inspired by the focal loss function~\cite{lin2017focal}, focuses on the actual background class by using a focal weight based on the maximum output from our backward background shift detector and hence ignores the dummy background detected that has high foreground probability. 
A  focal weight hyper-parameter \(\gamma\) smoothly adjusts the rate at which we down-weight the dummy background. 
Moreover, gradients do not propagate through the output of the backward background detector using stop gradients to avoid backpropagating the dummy background to our projector.
 Our proposed loss consists of two parts; the first one learns to distinguish between the actual background class and the foreground while avoiding the dummy background class. The second is used to disentangle new classes from everything else.

\begin{equation}\label{eq:bg_fg}
    \ell^{\theta^t}_{bg\_fg}(x^t,y^t) = -\frac{1}{|\set I|}\sum_{i \in\set I} (1 - \max_{\tilde{t} \in \{1:t-1\}} Fg^{\tilde{t}})^{\gamma}  \log \tilde{z}^t[i,y^t_i]
  \end{equation}
where \(\tilde{z}^t\) contains the probability of either being a background or a foreground
\begin{equation}
  \label{eq:bg_fg_cases}
  \tilde{z}^t[i,y^t_i] = \begin{cases}
    \displaystyle\sum_{k \in C^t } {z}^t[i,k]\;\;& \text{if}\ y^t_i \neq c_{bg} \\
    {z}^t[i,y^t_i]\;\;& \text{if}\ y^t_i=c_{bg}\,.
  \end{cases}
\end{equation}

Our intuition from \cref{eq:bg_fg} is that the focal loss from our backward background detector will guide the training to distinguish between the actual background and everything else without affecting old classes. Furthermore, to simplify the learning, we compare it against the probability of being any of the other classes as shown in \cref{eq:bg_fg_cases}.

The second part focuses on distinguishing new classes from everything else. We use the same unbiased cross-entropy loss~\cite{cermelli2020modeling}  to simultaneously compare the new classes' probabilities to the sum of the probabilities of background and old classes. Moreover, the first part of the loss learns the actual background, which avoids the confusion between old classes and the actual background when we use the unbiased cross-entropy loss~\cite{cermelli2020modeling} alone. 

\begin{equation} \label{eq:new_ce}
    \ell^{\theta^t}_{new}(x^t,y^t) = -\frac{1}{|\set I|}\sum_{i \in\set I}\log \hat{z}^t[i,y^t_i]\,,
 \end{equation}
  where \(\hat{z}^t\) represents the probability of being any of the old classes or the background \(\{c_{bg} \cup C^{1:t-1}\}\) versus new classes. 
\begin{equation}\label{eq:new_cases}
  \hat{z}^t[i,y^t_i] = \begin{cases}
      \displaystyle\sum_{k\in\mathcal{{C}}^{1:t-1}}{z}^t[i,k]\;\;& \text{if}\ y^t_i \in \{c_{bg} \cup C^{1:t-1}\}\\
      {z}^t[i,y^t_i]\;\;& \text{if}\ y^t_i \in \{C^t\}\,.
    \end{cases}
\end{equation}
Our loss function in subsequent steps used to learn new classes without pseudo-labeling becomes:
\begin{equation}\label{eq:total}
  \ell_{BACS}^{\theta^t} = \ell^{\theta^t}_{new}(x^t,y^t) + \ell^{\theta^t}_{bg\_fg}.
\end{equation}

Our intuition is that we can update the model to predict the new classes and simultaneously manage to distinguish between dummy backgrounds belonging to different steps and the actual background. 
It is worth noting that direct pseudo-labeling underperforms due to the erroneous high-confidence predictions~\cite{arazo2020pseudo} that will keep propagating throughout the training affecting the network's plasticity. 
For that reason~\cite{douillard2021plop} uses a threshold and an entropy loss to decide which pseudo-label to keep. 
On the other hand, our proposed strategy uses the backward background shift detector to distinguish foreground and background while ignoring the dummy background in the loss function.

\subsection{Decoder Multi-Classifier Initialization}

We now described how we add new classes to our decoder without having to initialize new classifiers as background. 
Existing \ac{CSS} methods initialize new classifiers as a background class incentivizing the network to predict new classes as background. 
This initialization trick hinders the convergence of the network to new classes and stabilizes the background's feature space at the expense of new classes. Furthermore, the multi-classifier setup tends to be specialized in the last set of classes, creating a bias towards new classes~\cite{hou2019learning}. 
To address these two drawbacks of the initialization trick, we use a transformer decoder inspired by the Segmenter architecture~\cite{strudel2021segmenter}, where a set of tokens represent our classes. 
First, we process class tokens, shown in colour in \cref{fig:framework}, jointly with patch embeddings, shown in grey in \cref{fig:framework}, through a transformer decoder layer. 
The output patches are passed through a dot-product operation to retrieve the final classification output. 
To add new classes, we append new class tokens to the input of the transformer decoder initialized as the average of all previous class tokens. 
Using class tokens makes adding new classes easier and reduces the bias toward new classes. Also, it improves the plasticity of our model with fewer parameters, as shown in \cref{fig:framework}.

\subsection{Catastrophic Forgetting}
In addition to the background shift and the initialization of classifiers, we tackle catastrophic forgetting by using a masked knowledge distillation and \acf{DER}~\cite{buzzega2020dark}. 
An effective method to tackle catastrophic forgetting is to set constraints using the previous step's model weights \(f_{\theta}^{t-1}\). 
These constraints enforce \(f_{\theta}^t\) to produce similar behavior on previously observed classes. 
A common constraint is to add soft knowledge distillation on the output layer~\cite{hinton2015distilling, li2017learning}. 
However, due to the forward background shift where the teacher \(f_{\theta}^{t-1}\) considers new classes as a background, a representation space drift occurs, affecting the stability of the network. For that purpose, unbiased knowledge distillation~\cite{cermelli2020modeling} compares the output of the teacher for the background class with either being a background or a new class in \(f_{\theta}^{t}\). 
Moreover, regularizing output-level knowledge distillation would reduce the plasticity of the network and force it to rely on a large number of initial classes. 
We propose a feature-based masked knowledge distillation to constrain the feature space of old classes without limiting the plasticity of the network. In that proposed knowledge distillation, we distill the knowledge of the penultimate layer  \(j^{t-1}\) for only a dummy background selected by our backward background detector. To summarize, our masked distillation loss with \(\delta\) a tuned threshold becomes:

\begin{equation} \label{eq:kd}
  \ell^{\theta^t}_{kd}(x^t,y^t) = (\max_{\tilde{t} \in \{1:t-1\}}  Fg^{\tilde{t}} > \delta) \cdot  \| (j^{t-1})^2 - (j^t)^2 \|.
\end{equation} 

Furthermore, we adapt dark experience replay~\cite{buzzega2020dark} to mitigate catastrophic forgetting while ignoring the background class as it might contain old or future classes. 
Following \cite{buzzega2021rethinking}, we use a balanced loss-aware reservoir sampling strategy. In brief, we make room for new examples by discarding less-critical samples in the buffer. The importance score is computed based on the number of samples belonging to that same class in the buffer and the loss value of the trained network denoting its difficulty. 

To summarize, our loss function becomes the sum of our background aware cross-entropy loss \( \ell_{BACS}^{\theta^t} \), experience replay \( \ell_{replay}^{\theta^t} \), dark knowledge replay \( \ell_{der}^{\theta^t} \), penultimate layer knowledge distillation \( \ell^{\theta^t}_{kd} \) and backward background detector training loss function \( \ell_{B}^{\omega^t} \).

\begin{equation} \label{eq:our_loss}
  \ell^{\theta^t} = \underbrace{ \ell_{BACS}^{\theta^t}}_\text{classification} +  \underbrace{ \alpha \ell_{der}^{\theta^t} + \beta \ell_{der++}^{\theta^t} + \kappa \ell^{\theta^t}_{kd}}_{ forgetting } + \ell_{B}^{\omega^t}
\end{equation}

In \cref{eq:our_loss}, we have a set of hyper-parameters \(\alpha\), \(\beta\), and \(\kappa\) tuned to guide the training to determine whether the focus is on stability or plasticity.  

%% file: sections/04-experiments.tex
\subsection{Setup and Datasets}
For a fair comparison, we use the same setup, hyper-parameters, and datasets as~\cite{michieli2019incremental}.
However, we propose testing on more challenging setups where we start from a few classes (e.g., 5, 2).

We evaluate our proposed approach on the most challenging overlap mode with forward and backward background shifts. For Pascal-VOC, we evaluate several dataset setups, e.g., $(15-1)$, which means we start our first task with 15 classes, then increment one class (for a total of 6 tasks). Furthermore, we evaluate more challenging setups, including $(10-1)$, $(5-3)$ and $(2-1)$. Similarly, Cityscapes $(14-1)$ means we start initially with 14 classes followed by increments of 1.

We evaluate our proposed method on two datasets: Pascal-VOC~\cite{everingham2015pascal} and Cityscapes~\cite{cordts2016cityscapes}. VOC contains an explicit background class and 20 classes, \(10,582\) training images, and \(1,449\) testing images. Cityscapes contain 19 classes and a set of unlabelled classes that we consider background taken from 21 cities. For all datasets, we use random resize, crop augmentation, and random horizontal flip during training following~\cite{douillard2021plop,michieli2019incremental}. The final image size for Pascal-VOC and Cityscapes is \( 512 \times 512\).
Hyper-parameters were tuned on a validation set created as a subset of the training set made of \(20\%\) of the images.

\subsubsection{Evaluation Metrics}

We use the mean Intersection-Over-Union (mIoU) as our evaluation metric. The IoU is defined as \(IoU = \frac{\text{true positive}}{\text{true positive + false negative + false positive}}\), which quantifies the accuracy of our method.  To evaluate our method's ability to preserve previously learned information, we calculate the mIoU for the classes of the initial task. Additionally,  We measure the IoU of newly added classes to denote the plasticity of the framework following~\cite{michieli2019incremental} metrics.

\begin{table*}[!ht]
  \centering
  \caption{ Experimental results on challenging setups of Pascal VOC 2012. \ac{BACS} outperforms existing baselines with a large margin on challenging setups where we start with a small number of classes and small increments.}
  \label{tab:sota_voc}
  \resizebox{\textwidth}{!}{%
    \begin{tblr}{
      row{even} = {c},
      row{3} = {c},
      row{5} = {c},
      row{7} = {c},
      row{9} = {c},
      row{11} = {c},
      cell{1}{2} = {c=3}{c},
      cell{1}{5} = {c=3}{c},
      cell{1}{8} = {c=3}{c},
      cell{1}{11} = {c=3}{c},
      cell{1}{14} = {c=3}{c},
      vline{2,5,8,11,14} = {2-11}{},
      hline{1-3,11-12} = {-}{},
        }
                 & VOC 10-1 (11 tasks) &               &               & VOC 15-1 (6 tasks) &               &               & VOC 5-3 (6 tasks) &               &               & VOC 5-1 (16 tasks) &               &               & VOC 2-1 (19 tasks) &               &                \\
      Method     & 0-10                & 11-20         & all           & 0-15               & 16-20         & all           & 0-5               & 6-20          & all           & 0-5                & 6-20          & all           & 0-2                & 3-20          & all            \\
      LwF-MC     & 4.65                & 5.90          & 4.95          & 6.40               & 8.40          & 6.90          & 20.91             & 36.67         & 24.66         & N/A                & N/A           & N/A           & N/A                & N/A           & N/A            \\
      ILT        & 7.15                & 3.67          & 5.50          & 8.75               & 7.99          & 8.56          & 22.51             & 31.66         & 29.04         & N/A                & N/A           & N/A           & N/A                & N/A           & N/A            \\
      MiB        & 12.25               & 13.09         & 12.65         & 34.22              & 13.50         & 29.29         & \textbf{52.2}     & 42.1          & 45.01         & 11.47              & 9.45          & 10.03         & 21.57              & 7.93          & 9.88           \\
      PLOP       & 44.03               & 15.51         & 30.45         & 65.12              & 21.11         & 54.64         & 17.48             & 19.16         & 18.68         & 0.12               & 9.0           & 6.46          & 0.01               & 5.22          & 4.47           \\
      PLOP + UCD & 42.3                & 28.3          & 35.3          & 66.3               & 21.5          & 55.1          & N/A               & N/A           & N/A           & N/A                & N/A           & N/A           & N/A                & N/A           & N/A            \\
      PLOPLong   & 61.06               & 18.56         & 40.83         & 72.06              & 26.66         & 61.2          & N/A               & N/A           & N/A           & N/A                & N/A           & N/A           & N/A                & N/A           & N/A            \\
      ReCALL     & 59.5                & \textbf{46.7} & \textbf{54.8} & 65.7               & \textbf{47.8} & 62.7          & N/A               & N/A           & N/A           & N/A                & N/A           & N/A           & N/A                & N/A           & N/A            \\
      RCIL       & 55.4                & 15.1          & 34.3          & 70.6               & 23.7          & 59.4          & N/A               & N/A           & N/A           & N/A                & N/A           & N/A           & N/A                & N/A           & N/A            \\
      BACS       & \textbf{64.1}       & 36.9          & 51.13         & \textbf{72}        & 44.32         & \textbf{65.4} & 46.61             & \textbf{44.8} & \textbf{45.3} & \textbf{35.2}      & \textbf{30.4} & \textbf{31.8} & \textbf{41.05}     & \textbf{29.7} & \textbf{31.35}
    \end{tblr}%
  }
\end{table*}

\subsubsection{Implementation Details}

In order to facilitate replication of our experimental setup, we have made available our implementation based on PyTorch Lightning, which can be accessed via a public GitHub repository\footnote{\url{https://github.com/mostafaelaraby/BACS-Continual-Semantic-Segmentation}}.

Following previous work~\cite{douillard2021plop,michieli2019incremental}, we use a Resnet-101 backbone~\cite{zhang2020resnest} pre-trained on ImageNet~\cite{he2015delving}, and instead of using DeepLab v3~\cite{yurtkulu2019semantic} decoding head we use segmenter~\cite{strudel2021segmenter} decoder with dynamic class tokens. We compared the number of parameters between our proposed decoder having \(55M\) parameters with DeepLab v3~\cite{yurtkulu2019semantic} decoder having \(58M\) parameters on Pascal VOC \(15-1\) final step showing the effectiveness of our decoder with fewer parameters.

We optimize the network for all baselines using SGD with an initial learning rate of \(10^{-2}\) reduced to \(10^{-3}\) in subsequent tasks with momentum \(0.9\). We use the same learning rate schedule, data augmentation, and class order as~\cite{michieli2019incremental,douillard2021plop}. For the memory size, we use \( \lvert \mathcal{M} \lvert = 300\) for both VOC and cityscapes. We train the network for \(30\) epochs with batch size \(24\) distributed on two GPUs for each task.

We select EWC~\cite{kirkpatrick2017overcoming}, LwF-MC~\cite{li2017learning}, and ILT~\cite{michieli2019incremental} to compare general results on VOC with results exacerbated from SSUL~\cite{cha2021ssul}. We compare specific CSS experiments against our re-implementation for both MiB~\cite{cermelli2020modeling}, PLOP~\cite{douillard2021plop}, and joint training as an upper bound.

\subsection{Quantitative and Qualitative Evaluation}

\subsubsection{Pascal VOC 2012}
Quantitative results in \cref{tab:sota_voc} compare our method with existing baselines on different scenarios, including the challenging \( {5-3, 10-1, 15-1}\) and the two tasks scenarios \( {15-5, 19-1}\) evaluated in the overlap mode having both backward and forward background shift.
We show that our proposed framework outperforms both PLOP~\cite{douillard2021plop} and MiB~\cite{michieli2019incremental} in terms of the plasticity on all tasks while retaining the knowledge acquired on old classes, especially in tasks starting with a small number of classes.
In both PLOP~\cite{douillard2021plop} and MiB~\cite{michieli2019incremental}, the pseudo-labeling and constrained teacher-student strategy limit the network's plasticity resulting in degraded performance on new classes.
In BACS, we mitigate this problem by discarding the pseudo-labeling strategy and simply relying on our backward background detector to solve the background shift problem.

\subsubsection{Cityscapes}

In \cref{tab:cityscapes_results}, we also evaluate our method with the challenging Cityscapes setup having 19 classes with the unlabelled classes folded into the background class.
We evaluate our method using a \(14-1\) setup. Here, BACS performs better than MiB and PLOP, specifically on new classes.

\begin{table}[htbp]
  \centering
  \captionsetup{justification=centering}
  \caption{Continual Semantic Segmentation results on Cityscapes 14-1 in Mean IoU (\%).}
  \label{tab:cityscapes_results}
  \small
  \begin{tabularx}{\columnwidth}{X|XX|c}
    \toprule
                    & \multicolumn{3}{c}{\textbf{14-1} (6 tasks)}                                  \\
    \textbf{Method} & 1-14                                        & 15-19          & \textit{all}  \\
    \midrule
    MiB             & \textbf{56}                                 & 15.7           & 46.5          \\
    PLOP            & 55.6                                        & 10             & 44.3          \\
    BACS            & 55.7                                        & \textbf{19.47} & \textbf{46.6} \\
    \bottomrule
  \end{tabularx}
\end{table}

\subsection{Impact of Class Ordering}
\begin{figure}[!h]
  \centering
  \includegraphics[width=0.65\columnwidth]{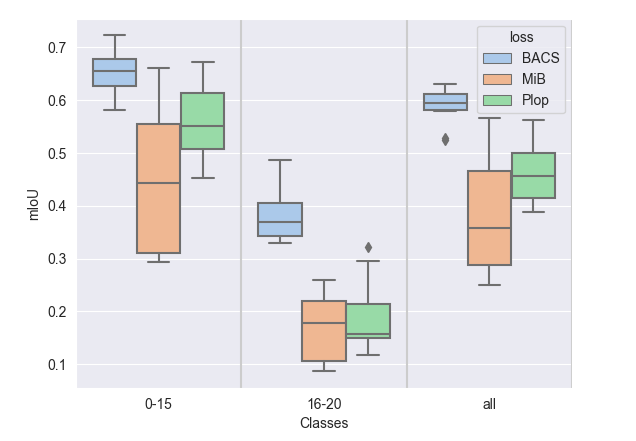}
  \caption{mIoU Evaluation of 10 different class orderings between BACS, MiB, and PLOP.}
  \label{fig:class_ordering}
\end{figure}

Here, we analyze the effect of class ordering on the performance of various \ac{CSS} frameworks. We perform experiments on ten different class orders on the challenging VOC \(15-1\) setup to analyze our robustness to the class order. In \cref{fig:class_ordering}, we display the mean and standard deviation of different class orders for various methods~\cite{douillard2021plop,cermelli2020modeling}. Experimental results show the robustness of our method to class order in the challenging overlap mode.

\begin{figure}[!htpb]
  \centering
  \includegraphics[width=\columnwidth]{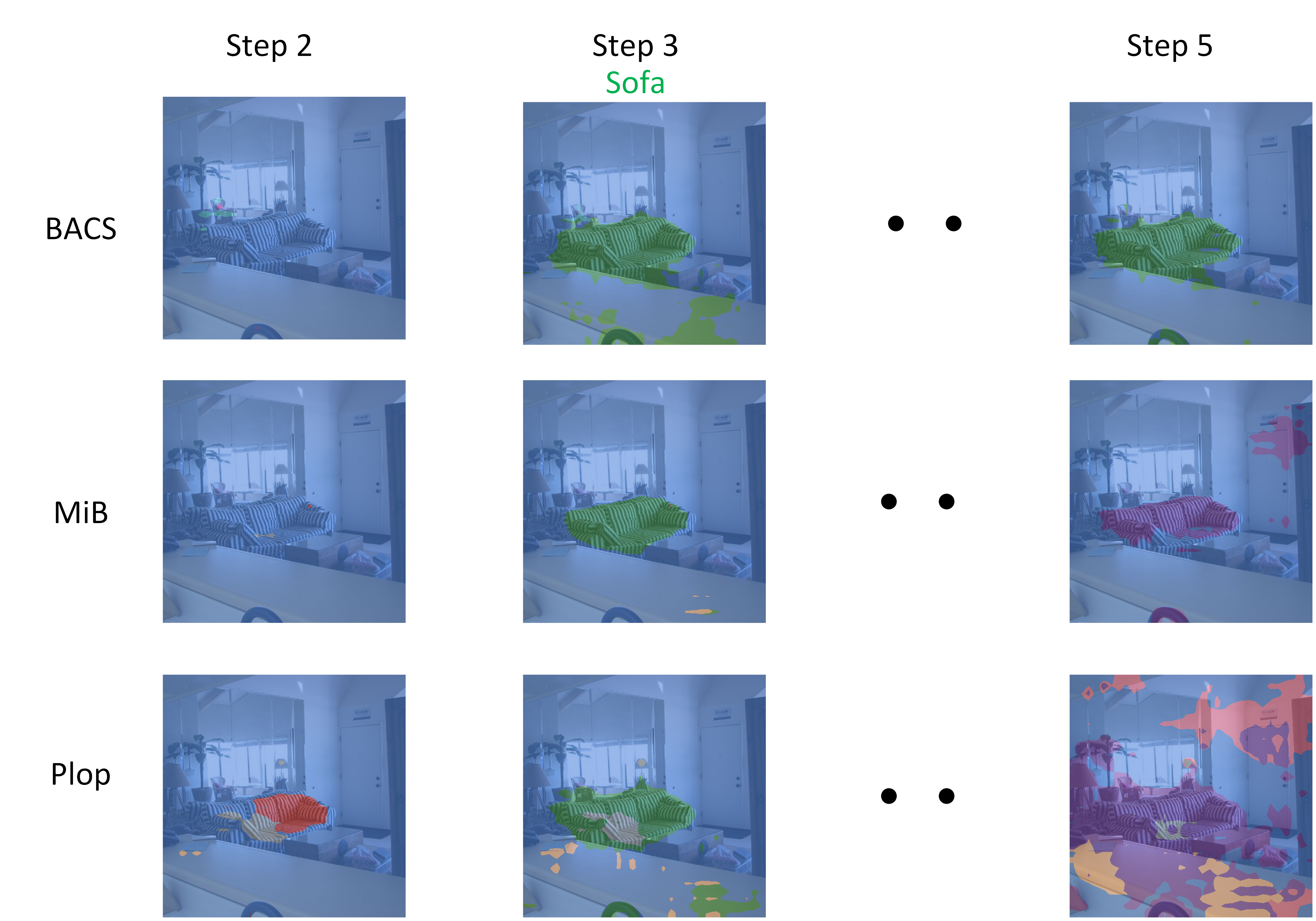}
  \caption{Qualitative comparison between \ac{BACS}, MiB and PLOP on \(15-1\) VOC setup. \textbf{Left column}: Predictions after learning two tasks, not including the upcoming sofa class. \textbf{Middle column}: Predictions after incrementing the sofa class. \textbf{Right column}:  Predictions at the end of the training.  }
  \label{fig:qualitative}
\end{figure}

\subsection{Ablations}
\begin{table}[htbp]
  \centering
  \captionsetup{justification=centering}
  \caption{Ablation study for BACS on VOC 15-1.}
  \label{tab:ablation_new}
  \small
  \begin{tabularx}{\columnwidth}{cccc|XXX}
    \toprule
    \multicolumn{4}{c|}{\textbf{Configurations}} & \multicolumn{3}{c}{\textbf{15-1 (6 tasks)}}                                                                         \\
    BACS                                         & DER                                         & MKD        & Dec        & 0-15        & 16-20         & all           \\
    \midrule
    \checkmark                                   & \checkmark                                  & \checkmark & \checkmark & \textbf{72} & \textbf{44.3} & \textbf{65.4} \\
    \checkmark                                   & \checkmark                                  & \xmark     & \checkmark & 68.8        & 29.7          & 59.5          \\
    \checkmark                                   & \checkmark                                  & \xmark     & \xmark     & 58.82       & 40.82         & 54.54         \\
    \checkmark                                   & \xmark                                      & \checkmark & \checkmark & 55          & 7             & 43.8          \\
    \bottomrule
  \end{tabularx}
\end{table}

Here, we analyze the effect of each proposed component of \ac{BACS} on VOC \(15-1\) in the overlap mode. \cref{tab:ablation_new} compares the results of each ablation case where we disable one or two of the components. The first row shows the result of \ac{BACS} with all the components, including dark experience replay (\ac{DER}), masked knowledge distillation(\ac{MKD}), and our transformer decoder (\ac{Dec}). When we disable \ac{MKD} in the second row, the performance on both old and new classes deteriorates, showing the benefits of soft knowledge distillation for both plasticity and stability. On the other hand, using the multi-classifier setup instead of \ac{Dec} improves the plasticity due to the increased number of parameters and fails to reduce the effect of the forward background shift due to the initialization trick used. In \ac{Dec}, we add a single parameter per new class, whereas the multi-classifier setup initializes a new \(1 \times 1\) convolutional head. Finally, in the last row, we show the effect of using \ac{DER}   on stabilizing the training regarding stability and plasticity.

\subsection{Initialization of New Tokens}

\begin{table}[!ht]
  \centering
  \caption{Initialization of new tokens corresponding to newly added classes. Random where we randomly initialize a new token per class, background means initializing new tokens as the background class and Mean, which averages all the old tokens to generate new ones.}
  \label{tab:new_token_init}
  \begin{tabularx}{\columnwidth}{l|XX|X}
    \toprule
                                  & \multicolumn{3}{c}{\textbf{VOC 15-1} (6 tasks)}                                 \\
    \textbf{Token Initialization} & 0-15                                            & 16-20         & \textit{all}  \\
    \midrule
    Random                        & 67.1                                            & 30            & 58.51         \\
    Background                    & 68.3                                            & 33.83         & 60            \\
    Mean                          & \textbf{72}                                     & \textbf{44.3} & \textbf{65.4} \\
    \bottomrule
  \end{tabularx}
\end{table}

In \cref{tab:new_token_init}, we show that \acf{Dec} is less sensitive to the way we initialize our tokens compared to the multi-classifier setup sensitivity discussed in~\cite{cermelli2020modeling}. In the first row, we show the effect of random initialization of new tokens, which slightly perturb the feature space due to the forward background shift. Next, we show the effect of initializing as a background similar to the initialization trick~\cite{cermelli2020modeling} but this time applied on a single parameter token. Finally, we compare our proposed initialization of tokens that reduces the feature space perturbation by simply taking the mean of all previous class tokens. We empirically show the effect of different token initialization on the stability of the network throughout the training.

%% file: sections/05-conclusion.tex
In this work, We proposed a new framework \acf{BACS} to address three principal challenges in \acf{CSS}: catastrophic forgetting, background shift, and the initialization of new class heads.  We proposed four main contributions, including a backward background shift detector, a variation of cross-entropy loss, a masked feature-based \acf{MKD} and finally, a transformer decoder. 
These main contributions demonstrate significant improvements in handling a large number of tasks, particularly when starting from a small set of classes. Our experiments confirm that each component contributes substantially to the overall effectiveness of the model, outperforming existing methods across standard benchmarks.